\pdfoutput=1

\documentclass[11pt]{article}

\usepackage[final]{acl}

\usepackage{times}
\usepackage{latexsym}

\usepackage[T1]{fontenc}

\usepackage[utf8]{inputenc}

\usepackage{microtype}

\usepackage{inconsolata}

\usepackage{graphicx}
\usepackage{amsmath}
\usepackage{amssymb}  
\usepackage{graphicx}
\usepackage{xspace}
\usepackage{makecell}
\usepackage{adjustbox}
\usepackage{comment}
\usepackage{booktabs}
\usepackage{multirow}
\usepackage{caption}
\usepackage{subcaption,comment}
\newcommand\tf[1]{\textbf{#1}}

\newcommand{\model}{\textsc{DynRank}\xspace}

\title{DynRank: Improving Passage Retrieval with Dynamic
Zero-Shot Prompting Based on Question Classification}

\author{
    \textbf{Abdelrahman Abdallah, Jamshid Mozafari, Bhawna Piryani,} \\
    \textbf{Mohammed M.Abdelgwad, Adam Jatowt} \\
    University of Innsbruck \\
    \texttt{\{abdelrahman.abdallah, jamshid.mozafari, bhawna.piryani,} \\
    \texttt{mohammed.ali, adam.jatowt\}@uibk.ac.at}
}

\begin{document}
\maketitle
\begin{abstract}
This paper presents \model, a novel framework for enhancing passage retrieval in open-domain question-answering systems through dynamic zero-shot question classification. Traditional approaches rely on static prompts and pre-defined templates, which may limit model adaptability across different questions and contexts. In contrast, \model introduces a dynamic prompting mechanism, leveraging a pre-trained question classification model that categorizes questions into fine-grained types. Based on these classifications, contextually relevant prompts are generated, enabling more effective passage retrieval. We integrate \model into existing retrieval frameworks and conduct extensive experiments on multiple QA benchmark datasets.

\end{abstract}

\section{Introduction}
Document retrieval plays a crucial role in many NLP tasks, particularly in open-domain question-answering (ODQA) systems~\cite{gruber2024complextempqa,abdallah2023exploring,piryani-etal-2024-detecting}. In ODQA, a passage is first retrieved and then processed to answer a question. In these systems, the retriever component initially retrieves the most relevant passages from document resources like Wikipedia relevant to the posed question. Subsequently, the reader component examines the retrieved passages to detect an answer. This pipeline causes that the effectiveness of ODQA systems heavily relies on the quality and efficiency of the retriever. Recent advancements in NLP have focused on enhancing the performance of the retriever component by reranking the retrieved passages to improve the likelihood of retrieving the most relevant content and appear on the top~\cite{,mozafari2024exploring,abdallah2023generator,abdallah2024arabicaqa,pradeep2023rankvicuna,sachan2022improving,Sun2023IsCG,zhuang2024setwise,pradeep2023rankzephyr}.

 \begin{figure}[t!]
\centering
\includegraphics[width=0.35\textwidth]{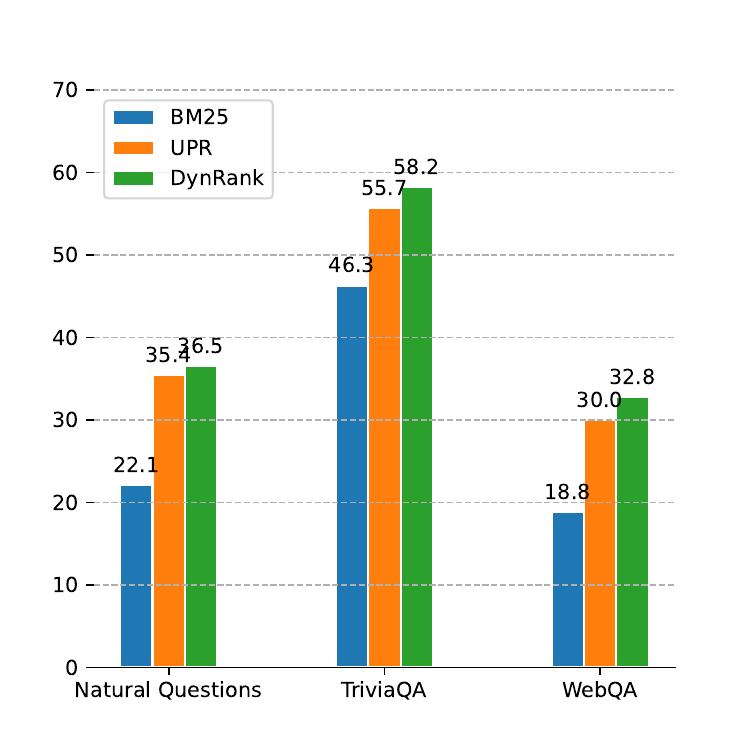}
\caption{
    Top-1 Accuracy after re-ranking the top 1,000 passages retrieved by BM25 with \model comparing with UPR \cite{sachan2022improving} on the Natural Questions  TriviaQA and WebQA datasets.
}
\label{fig:intro}
\end{figure}
With the development of large language models (LLMs), many researchers focused on employing LLMs to rerank retrieved passages using diverse strategies and prompting techniques~\cite{Sun2023IsCG}. Despite the success of LLM-based reranking methods such as UPR~\cite{sachan2022improving} which generate a question from on each retrieved passage and then re-rank the passages based on how similar their generated questions are to the original question, one significant limitation remains. These models often rely on static and pre-defined instructions during reranking, which can reduce their adaptability across diverse queries and contexts. This static nature restricts the dynamic interaction between the query and retrieved passages, potentially leaving relevant information unused. While methods like RankGPT~\cite{Sun2023IsCG} have shown performance by generating permutations of passages based on relevance, they still follow a static instruction approach. RankGPT excels in zero-shot passage re-ranking tasks by employing a fixed set of instructions, but it does not dynamically adjust based on the query context. In contrast, our approach aims to overcome this limitation by incorporating more dynamic interactions between the queries and retrieved passages. Similar to UPR~\cite{sachan2022improving} it generates questions from passages to be used for passage reranking; yet this generation is conditioned by the inferred categories of a target question. A detailed discussion of related works is provided in Appendix~\ref{appendix:related_work}.


In response to the above-mentioned challenges, we propose DynRank, a novel re-ranking framework designed to dynamically generate prompts tailored to each specific question. 
We leverage methods from the Question Classification (QC) task~\cite{cortes-etal-2020-empirical}, which categorizes questions into two groups: coarse-grained, indicating the major question type, and fine-grained, specifying the minor question type. Table~\ref{tab:major_minor_categories} in Appenidx~\ref{appendix:categories} presents the question types in detail.
By incorporating a fine-grained question classification model, DynRank adapts to the context of each query, generating prompts that are more relevant and contextually appropriate. 
Our contributions are threefold: 

\begin{enumerate} 
\vspace{-.2cm}

\item We introduce a dynamic prompting mechanism that leverages question classification to tailor prompts based on the specific context of each query. 
\vspace{-.3cm}
\item We integrate DynRank into existing retrieval framework and demonstrate its compatibility and ease of implementation. 
\vspace{-.3cm}
\item We conduct extensive experiments on benchmark QA datasets, demonstrating that 
DynRank outperforms both traditional and state-of-the-art re-ranking methods.
\vspace{-.2cm}

\end{enumerate}

\begin{figure*}[t]
  \centering
  \includegraphics[width=1.6\columnwidth]{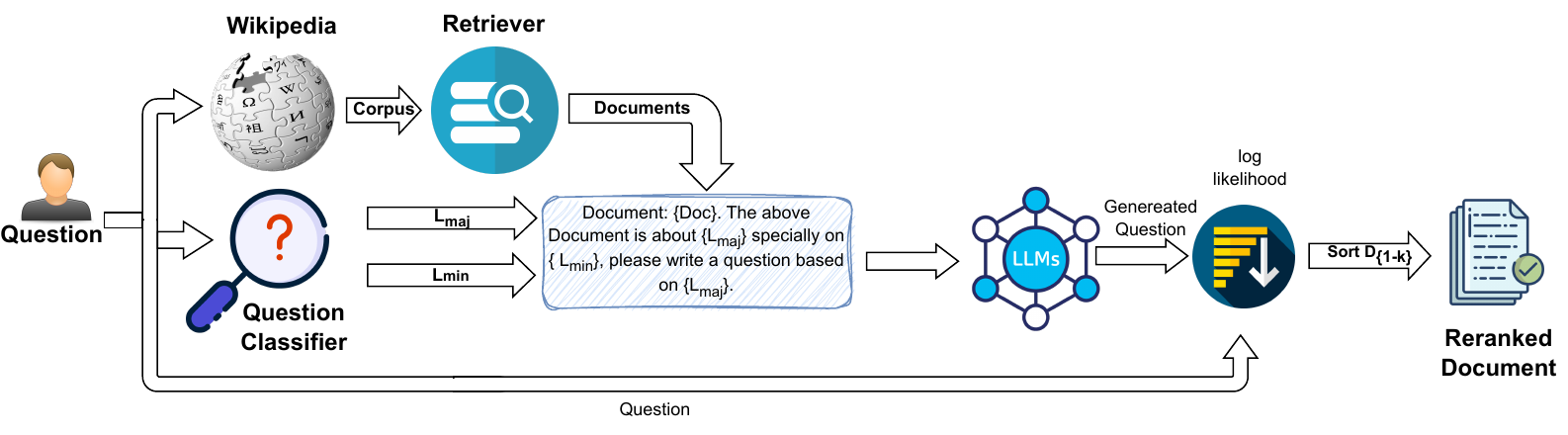}
\caption{The architecture of the proposed \model framework. First, retrieve relevant documents a retriever. The retrieved documents are then re-ranked based on a dynamically generated prompt, tailored to the question's classification into major ($L_{maj}$) and minor ($L_{min}$) types. A large language model (LLM) processes this prompt to re-rank the top-k documents, improving retrieval accuracy. }
  \label{fig:reranking-method}
\end{figure*}
\vspace{-.2cm}

\section{Method}
\label{sec:method}

Figure~\ref{fig:reranking-method} presents an overview of our approach for open-domain retrieval, introducing a novel dynamic zero-shot question generation method (\model) for re-ranking passages retrieved by any existing retriever.

\subsection{Retriever}
\label{sec:method-retriever}

Let $\mathcal{D} = \{\boldsymbol{d}_1, \ldots, \boldsymbol{d}_M \}$ be a collection of evidence documents. Given a question $\boldsymbol{q}$, the retriever selects a subset of relevant passages $\mathcal{Z} \subset \mathcal{D}$, one or more of which ideally contains the answer to $\boldsymbol{q}$. Our method works with passages obtained from any retriever — either based on sparse representations like BM25 or dense representations like DPR. We assume that the retriever provides the $K$ most relevant passages, denoted as $\mathcal{Z} = \{\boldsymbol{z}_1, \ldots, \boldsymbol{z}_K\}$. 

\subsection{Question Classification}
\label{sec:method-classification}

We fine-tune the RoBERTa model~\cite{2019arXiv190711692L} on the UIUC dataset~\cite{li-roth-2002-learning} to build a question classifier. This fine-tuned model categorizes questions based on different levels of granularity, including 5 coarse-grained classes and 50 fine-grained classes.
Given an input question $\boldsymbol{q}$, the question classification model assigns it to a major type $l_{maj}$ and a minor subtype $l_{min}$. Formally, let $Q$ be the set of all possible questions, and $L_{maj}$ and $L_{min}$ be the sets of major and minor types, respectively. The classification function $\mathcal{C}: Q \rightarrow L_{maj} \times L_{min}$ maps each question to its corresponding type and subtype:

\begin{equation}
(l_{maj}, l_{min}) = \mathcal{C}(\boldsymbol{q})
\end{equation}

\subsection{Dynamic Prompt Generation}
\label{sec:method-prompt}

The dynamic prompt generator constructs a prompt $p$ tailored to each question based on its classification result $(l_{maj}, l_{min})$. Let $\mathcal{T}$ be a template function that generates prompts. The prompt $p$ is generated as follows:   

\begin{equation}
p = \mathcal{T}(l_{maj}, l_{min})
\end{equation}

For instance, Given the major type $l_{maj}$ as "human" and the minor type $l_{min}$ as "individual," the generated prompt for the passage could be: \textit{Document: [passage]. The above Document is about humans, specially
on individuals, please write a question
based on humans. }


\subsection{Re-ranking with Pre-trained Language Models}
\label{sec:method-reranking}

Given the dynamically generated prompt $p$ and a set of retrieved passages $\mathcal{Z} = \{\boldsymbol{z}_1, \ldots, \boldsymbol{z}_K\}$ for the question $\boldsymbol{q}$, the goal of the re-ranking module is to reorder the passages such that the ones containing the correct answer are ranked higher. We define the relevance score $s_i$ for each passage $\boldsymbol{z}_i$ as the log-likelihood of generating the question $\boldsymbol{q}$ given $\boldsymbol{z}_i$ and $p$. We estimate the relevance score $s_i$ as the conditional probability of the question $\boldsymbol{q}$ given the passage $\boldsymbol{z}_i$ and the prompt $p$ using a pre-trained language model. The scoring function $\mathcal{S}$ is defined as:

\begin{equation}
s_i = \mathcal{S}(\boldsymbol{q} \mid \boldsymbol{z}_i, p)
\end{equation}


The log-likelihood $\log P(\boldsymbol{q} \mid \boldsymbol{z}_i, p)$ is then computed as:

\begin{equation}
\log P(\boldsymbol{q} \mid \boldsymbol{z}_i, p) = \frac{1}{|\boldsymbol{q}|}\sum_{t=1}^{|\boldsymbol{q}|} \log P(q_t \mid \boldsymbol{q}_{<t}, \boldsymbol{z}_i, p; \Theta)\end{equation}

where $q_t$ is the $t$-th token of the question $\boldsymbol{q}$, $\boldsymbol{q}_{<t}$ represents all tokens before $q_t$, and $\Theta$ denotes the parameters of the pre-trained language model. The passages are then re-ranked based on their computed scores $s_i$. The re-ranking function $\mathcal{R}$ sorts the passages by descending order of $s_i$:

\begin{equation}
\mathcal{R}(\mathcal{Z}) = \text{sort}(\mathcal{Z}, \text{by } s_i)
\end{equation}

This approach ensures that the most relevant passages, as determined by the dynamically generated prompts, are ranked higher, thereby improving the accuracy of passage retrieval in open-domain question-answering systems.

\begin{table*}[t!]
    \centering
    \begin{adjustbox}{width={0.7\textwidth}}
    \begin{tabular}{@{}l | c c c c | c c c c | c c c c  @{}} 
    \toprule
    \tf{Retriever} & \multicolumn{4}{c|}{\tf{NQ }} & \multicolumn{4}{c|}{\tf{TriviaQA}} & \multicolumn{4}{c}{\tf{WebQ}}  \\ 
                        & Top-1  & Top-20 & Top-100 & Avg & Top-1 & Top-20 & Top-100 & Avg  & Top-10& Top-20 & Top-100 & Avg \\
    \midrule
    \multicolumn{13}{c}{\textit{Unsupervised Retrievers}} \\
    \midrule

    BM25              & 22.1 & 62.9 & 78.2 & 54.4 &  46.3 & 76.4 & 83.1 &  68.6 & 18.8  & 62.4 &75.4 & 52.2  \\

    BM25 + UPR        &  35.4  &78.4 & 85.2 & 66.3 & 55.7  & 82.8 & 86.4 & 74.9 & 30.0  &72.8 &81.6 &61.4 \\

    BM25 + \model   &  \textbf{36.5} & \textbf{78.7} & \textbf{85.5} &  \textbf{66.9} &  \textbf{58.2}  & \textbf{83.1} & \textbf{86.6} & \textbf{75.9} &  \textbf{32.8} & \textbf{73.8}& \textbf{81.9}& \textbf{62.8} \\

    \midrule
    Contriever        &  22.1 & 67.8 & 80.5 & 56.8 & 34.1 & 73.9 & 82.9 & 63.6 &  19.9 & 65.6& 80.1 &  55.2 \\

    Contriever + UPR  & 36.3 & 79.6 &86.6 & 67.5 &  56.7  & 82.8  & 86.4 & 75.3&   30.0 &  74.6 & 82.9 &62.5 \\

    Contriever + \model   &   \textbf{37.0}  &  \textbf{80.1}& \textbf{86.9} & \textbf{68.0 }& \textbf{58.5}  &\textbf{82.7} &\textbf{86.4}  &\textbf{75.8} & \textbf{32.8} & \textbf{75.1} &\textbf{83.4 }& \textbf{63.7}\\

    \midrule
    \multicolumn{13}{c}{\textit{Supervised Retrievers}} \\
    \midrule
    DPR               &  48.6  & 79.1 &85.7 &  71.1 & 57.4  & 79.7 & 85.0& 74.0  & 44.8  & 74.6 & 81.6 & 67.0\\

    DPR + UPR         &  42.5  & 83.3 & 88.5 & 71.4 & 61.3   & 84.2 & 87.2&  77.5 &  34.9  & 77.1 & 83.8 &   62.2\\ 
    
    DPR + \model    &   \textbf{42.5} & \textbf{83.9} & \textbf{89.0} & \textbf{71.8}& \textbf{61.9} & \textbf{85.2} & \textbf{88.1} &  \textbf{78.4}  & \textbf{37.2} & \textbf{77.7}   &  \textbf{84.0} &  \textbf{66.3} \\

    \bottomrule
    \end{tabular}
    \end{adjustbox}
    \caption{
    Top-\{1, 20, 100\} retrieval accuracy on the test set of datasets before and after re-ranking the top 1000 retrieved passages.
    }
    \label{tab-open-domain-rank-llama7b}
    \end{table*}
\section{Experiments}

\subsection{Datasets}
Our experiments utilize several benchmark datasets including Natural Questions (NQ)~\cite{Kwiatkowski2019natural}, TriviaQA~\cite{joshi2017triviaqa}, WebQuestions (WebQ)~\cite{berant-etal-2013-semantic}, the BEIR benchmark~\cite{thakur2021beir}, and the Question Classification dataset~\cite{li-roth-2002-learning}. These datasets are crucial for evaluating the performance of question-answering systems across different types of queries and tasks. For detailed descriptions and statistics see Appendix~\ref{appendix:datasets}.
\subsection{Experimental Setup}

\paragraph{Open-Domain QA:}
For ODQA tasks, we evaluate our method using various retrievers including BM25, Contriever, and DPR. We follow the approach described in~\cite{sachan2022improving} to retrieve and re-rank the top-1,000 passages, reporting results in terms of top-\{1,20,100\} retrieval accuracy.

\paragraph{BEIR:}
For the BEIR datasets, we re-rank the top 100 passages retrieved by BM25 using the Pyserini toolkit~\cite{lin2021pyserini}, assessing the effectiveness of our method using the nDCG@10 metric.

\paragraph{Implementation Details:}
We employ open-source implementations for all retrievers, using the Pyserini toolkit~\cite{lin2021pyserini} for BM25 and implementations from~\cite{singh2021end} for MSS and DPR. Retrievers for the Contriever are sourced from Huggingface checkpoints~\cite{wolf2020transformers}. Experiments are conducted on NVIDIA A40 GPUs within a high-performance computing cluster. Detailed configuration parameters and settings are provided in Appendix~\ref{appendix:implementation_details} and~\ref{appendix:hyperparameters_asd}.

\section{Results}

\begin{table}
    \centering
    \begin{adjustbox}{width={0.7\columnwidth}}
        \begin{tabular}{@{}lcc@{}}
        \toprule
        \multirow{2}{*}{Model} & \multicolumn{2}{c}{Accuracy}  \\
                               & Coarse-grained        & Fine-grained          \\ \midrule
        Bert-base-cased        & 97.2          & \textbf{91.8} \\
        Bert-large-cased       & 98.0            & 83.8          \\
        DistilBert-base-cased  & 97.0            & 88.2          \\
        Albert-base-v2         & 96.0            & 87.8          \\
        Albert-large-v2        & 95.8          & 79.2          \\
        RobertA-base           & 97.2          & 90.6          \\
        RobertA-large          & \textbf{97.8} & 89.4          \\ \bottomrule
        \end{tabular}
    \end{adjustbox}
    \caption{The accuracy of question classifier based on the various models.}
  \label{tab:qc_accuracy}
\end{table}
\paragraph{Question Classification}
We evaluated several models as the question classifier, including BERT~\cite{devlin-etal-2019-bert}, DistilBERT~\cite{2019arXiv191001108S}, ALBERT~\cite{an2020ALBERT}, and RoBERTa~\cite{2019arXiv190711692L}. Table~\ref{tab:qc_accuracy} shows that while RoBERTa excelled in coarse-grained classification, BERT-base performed optimally in fine-grained scenarios, underscoring the strengths of different models depending on the granularity of the classification task.

\paragraph{Open-Domain QA}
As shown in Table~\ref{tab-open-domain-rank-llama7b}, \model improves retrieval performance across various retrievers and datasets. When applied to BM25, \model increased top-20 accuracy by 3.3\% on NQ and 3.6\% on TriviaQA, and it achieved a 2.7\% improvement on average across all datasets with Contriever. These results demonstrate \model's effectiveness in enhancing performance, even when working with unsupervised models. The supervised DPR model also benefited from \model, with a 4.4\% boost in top-10 accuracy and a 4.8\% increase in top-20 accuracy on the NQ dataset, along with a 4.5\% improvement in top-1 accuracy on TriviaQA. Compared to UPR, a recent unsupervised re-ranking method, \model consistently outperformed it, including a 1.4\% higher top-10 accuracy on WebQuestions. These gains highlight \model’s ability to dynamically tailor prompts, resulting in more effective retrieval than static approaches like UPR.

\vspace{-0.2cm}
\paragraph{BEIR Benchmark}

Table~\ref{table:benchmark} presents the nDCG@10 results for four BEIR datasets. \model shows substantial improvements over the BM25 baseline, achieving an average nDCG@10 score of 54.1 across the BEIR datasets, compared to 47.7 with BM25. Specifically, \model outperforms BM25 by 6.0\% on the Covid dataset and by 4.5\% on SciFact. 
Compared to other state-of-the-art methods such as RankGPT, \model achieves competitive results. While RankGPT slightly outperforms \model on Covid and SciFact, \model shows stronger performance on NFCorpus and Signal, highlighting its versatility and effectiveness across various domains.

 \begin{figure}[t!]
\centering
\includegraphics[width=0.35\textwidth]{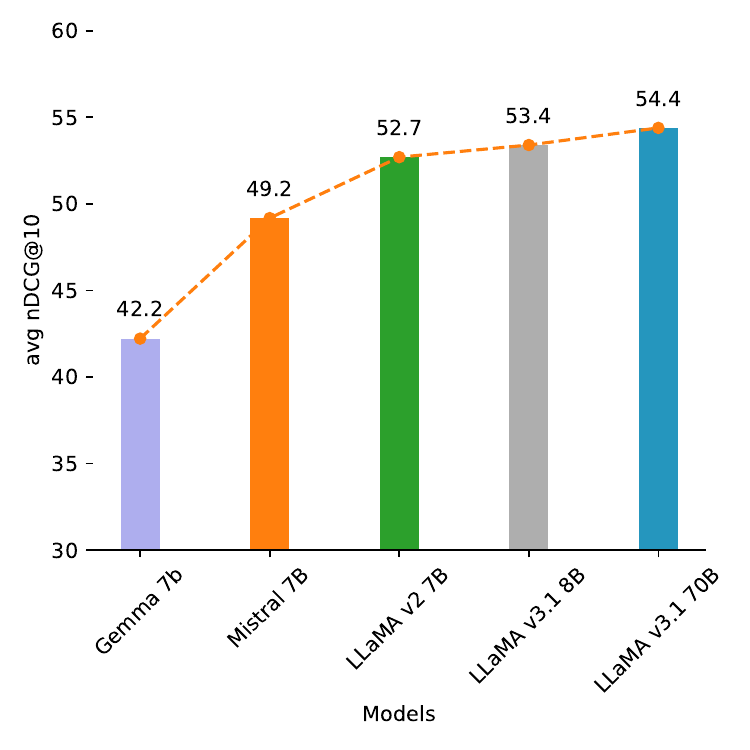}
\caption{Average nDCG@10 scores for different language models on the BEIR benchmark.
}
\label{fig:com}
\end{figure}
\vspace{-0.2cm}
\paragraph{Impact of Different Language Models}

To assess the impact of different language models on retrieval performance, we evaluated several pre-trained models on the BEIR benchmark, specifically measuring nDCG@10 across different datasets. The models compared in this analysis include Gemma 7B, Mistral 7B, LLaMA v2 7B, LLaMA v3.1 8B, and LLaMA v3.1 70B.  Fig \ref{fig:com} presents the average nDCG@10 scores for each model. Our findings show that LLaMA v3.1 70B achieved the highest performance with an nDCG@10 of 54.39, followed closely by LLaMA v3.1 8B with a score of 53.4. Both outperformed  Mistral 7B and LLaMA v2 7B, which achieved scores of 49.18 and 52.7.

\begin{table}[!t]
\centering
\setlength\tabcolsep{3pt}
\begin{adjustbox}{width={0.5\textwidth}}
\begin{tabular}{l  cccccccc | c }

\toprule
\textbf{Method}  & Covid &  NFCorpus   & SciFact &  Signal  & BEIR (Avg) \\

\midrule

BM25
 & 59.47 & 30.75  & 67.89 & 33.05 & 47.7 \\

\midrule
UPR~\cite{sachan2022improving}  & 68.11 & 35.04 & 72.69 & 31.91  & 51.9
\\

 SGPT-CE~\cite{muennighoff2022sgpt}  & 23.4 & 37.0 & 46.6 & 48.0 & 38.7\\

 HyDE~\cite{gao2022precise}  & 27.3 & - &  46.6 & - &- \\



RankGPT~\cite{Sun2023IsCG} & 76.67 & 35.62  & 70.43 & 32.12 & 53.7 
\\

\midrule
\model & 76.06 & 35.80  & 72.80 &32.93 & \textbf{54.39} \\

\bottomrule
\end{tabular}
    \end{adjustbox}

\caption{Results (nDCG@10) on  BEIR.}
\label{table:benchmark}

\end{table}

\section{Conclusion}

We introduced \model, a framework that improves open-domain question answering by generating tailored prompts using a question classification model. Experiments across multiple datasets demonstrate that \model outperforms static prompt methods, significantly enhancing retrieval performance and accuracy when integrated with both supervised and unsupervised retrievers.

\section{Limitations}

Despite the promising results, \model has several limitations that warrant further investigation:

\begin{itemize}
    \item \textbf{Computational Complexity:} The dynamic generation of prompts, while beneficial for accuracy, introduces additional computational overhead. 
    \item \textbf{Dependence on Pre-trained Models:} \model's performance is heavily dependent on the quality and size of the pre-trained language models used for generating the question based on the dynamic prompt.
\end{itemize}
\bibliography{custom}

\appendix
\section{Dataset Details}
\label{appendix:datasets}

This appendix provides comprehensive information about the datasets used in our experiments. Each dataset is chosen for its relevance to the field of natural language processing, providing a robust framework to evaluate the effectiveness of our proposed methods.

\subsection*{Natural Questions (NQ)}

The Natural Questions dataset contains 79,168 training examples, 8,757 development examples, and 3,610 test examples. It comprises questions derived from Google searches, paired with Wikipedia pages that provide the context necessary for answering the questions.

\subsection*{TriviaQA}

TriviaQA consists of 78,785 training examples, 8,837 development examples, and 11,313 test examples. It is made up of trivia questions sourced from trivia and quiz-league websites, annotated with evidence documents that support answers.

\subsection*{WebQuestions (WebQ)}

WebQuestions features 5,810 question-answer pairs, based on Freebase data, developed to facilitate research in question answering using structured knowledge sources.

\subsection*{BEIR Benchmark}

The BEIR (Benchmark for Evaluating Information Retrieval) benchmark is a diverse set of retrieval tasks and domains that assess the generalization capabilities of retrieval systems. It includes several datasets covering various retrieval tasks such as fact-checking, question answering, and others.

\subsection*{Question Classification Dataset}

The Question Classification dataset is used for training models to categorize questions into coarse and fine-grained classes, enhancing the targeting of retrieval mechanisms. It consists of questions labeled according to their type, which helps in tuning retrieval systems to the specific needs of the question being asked.
\begin{table}[htb]
	\centering
	
	\begin{tabular}{@{}l|lll@{}}
		\toprule Dataset  & Train  & Dev   & Test   \\
		\midrule 
        TriviaQA & 78,785 & 8,837 & 11,313 \\
		NQ       & 79,168 & 8,757 & 3,610  \\
		WebQ     & 3,417  & 361   & 2,032  \\
		\bottomrule
	\end{tabular}
 	\caption{Statistics of the ODQA datasets: TriviaQA, NQ, and WebQ.}
\label{tbl:dataset_statistics}
\end{table}

\section{Implementation Details}
\label{appendix:implementation_details}

This appendix provides comprehensive implementation details for our experiments. We aim to ensure reproducibility and provide insights into the specific configurations used across different setups.

\subsection*{Open-Domain QA}
\paragraph{Retrievers:}
We use BM25, Contriever, and DPR for initial retrieval. For BM25, the Pyserini toolkit is utilized, while for MSS and DPR, we employ implementations from Singh et al.~\cite{singh2021end}. Contriever utilizes checkpoints from Huggingface~\cite{wolf2020transformers}.

\paragraph{Re-Ranking:}
The top-1,000 passages retrieved are re-ranked according to our method, and results are reported for top-\{1,20,100\} retrieval accuracy.

\subsection*{BEIR Benchmark}
\paragraph{Re-Ranking:}
For the BEIR benchmark, the top 100 passages retrieved using BM25 are re-ranked using our methodology. The nDCG@10 metric is used to assess performance.

\subsection*{Hardware:}
All experiments are conducted using NVIDIA A40 GPUs on a high-performance computing cluster, ensuring significant computational resources.

\subsection*{Question Classification Model Training:}
\paragraph{Training Details:}
For Question Classification, we use the Transformers library~\cite{wolf-etal-2020-transformers} to fine-tune models. Training is conducted over 5 epochs with a batch size of 64 and a dropout rate of 0.1. A learning rate of 2e-5 is applied for base models, while 5e-6 is used for large models. This fine-tuning is essential for enhancing the model's ability to accurately classify questions into coarse and fine-grained categories.

\section{Hyperparameters}
\label{appendix:hyperparameters_asd}

For our experiments in open-domain question answering (ODQA) and BEIR retrieval tasks, we employed various large language models (LLMs) and configurations tailored to each dataset.

For the open-domain question answering tasks, we utilized the T0-3B model from Hugging Face, which is known for its ability to perform zero-shot learning based on prompts. Tokenized input passages were at 512 tokens and the generated question was 128 tokens.  The top 1,000 passages retrieved were re-ranked for each query.
For the BEIR benchmark, which involves a diverse set of retrieval tasks, we used larger models to accommodate the complexity of the retrieval tasks. The following models and settings were employed:
\begin{itemize}
    \item \textbf{Models:} LLaMA v3.1 (70B, 8B), LLaMA v2 7B, Mistral 7B, and Gemma 7B.
    \item \textbf{Batch Size:} 16 per GPU.
    \item \textbf{Evaluation Metric:} The primary evaluation metric used was nDCG@10, a common metric for evaluating the relevance of ranked lists in retrieval tasks.
\end{itemize}

These configurations were chosen to balance retrieval performance, model size, and computational efficiency, ensuring that our approach was scalable across different datasets.

\section{Related Work}
\label{appendix:related_work}
Passage re-ranking in question answering (QA) systems has been a focus of considerable research, as it significantly impacts the effectiveness of retrieving the most relevant information from a large corpus. Traditional approaches often rely on unsupervised methods like BM25 and dense retrievers such as DPR, which have laid the foundation for initial retrieval steps~\cite{wu2024passage}.

Recent advancements have shifted towards leveraging large pre-trained language models (LLMs) like BERT for re-ranking. Nogueira and Cho's work on BERT-based re-ranking demonstrated substantial improvements in passage retrieval accuracy by fine-tuning BERT on retrieval-specific tasks, showing that neural models could outperform traditional retrieval methods~\cite{nogueira2019passage,padigela2019investigating}. This approach set a new benchmark, particularly in datasets such as MS MARCO and TREC-CAR, highlighting the effectiveness of transformer-based architectures in understanding query-passage relevance.

Question Classification is a task that requires determining the type of answer for a given question. Given the task's dependency on the language of the questions, the most effective methods leverage external linguistic resources~\cite{tayyar-madabushi-lee-2016-high, iet-sen.2018.0006, 10.1007/978-981-10-5687-1_54}. However, some methods have been developed to address this task without relying on the language of the questions~\cite{YANG2019247, 2015arXiv151108630Z, 8622331}. In this study, we adopt the first approach, as we focus on English, the most widely used language.

Further innovations include techniques like Passage-specific Prompt Tuning (PSPT), which fine-tunes a small number of parameters while keeping the core LLM parameters fixed. PSPT dynamically adjusts prompts based on individual passages, enhancing the model's adaptability across various contexts and improving re-ranking performance by incorporating passage-specific knowledge~\cite{wu2024passage}. This method stands out for its ability to tailor the retrieval process more closely to the nuances of the input data. Another significant contribution is the work on RankQA, which extends the conventional two-stage QA process with an additional re-ranking stage. This method integrates retrieval and comprehension features to re-rank answers, thereby improving the accuracy and robustness of QA systems. RankQA's simple yet effective design underscores the importance of combining multiple features for enhanced passage retrieval~\cite{kratzwald2019rankqa}. Moreover, generative models have also been explored for passage retrieval. These models, while computationally intensive, show promise by generating candidate passages that are then ranked for relevance. However, their high computational cost remains a challenge for practical deployment in large-scale systems~\cite{nagumothu2023semantic}. Our proposed method, Dynamic Zero-Shot Question Generation for Re-ranking (\model), builds on these advancements by introducing dynamic prompt generation tailored to each question. By classifying questions into fine-grained categories and generating contextually relevant prompts, \model aims to enhance the adaptability and accuracy of passage retrieval, addressing some of the limitations of static prompt-based approaches.
\section{Case Study}
The case study section presents a comparison between questions generated by two re-ranking methods: UPR and \model. The tables showcase how each model generates questions from passages first retrieved by BM25. The examples demonstrate clear differences between UPR and \model. UPR tends to generate more generic or passage-level questions, while \model generates more specific and contextually aligned questions. This highlights the advantage of the \model in producing more relevant and detailed queries, which are better suited for open-domain question-answering tasks. The tables help illustrate the practical impact of these models by showcasing the real output from the datasets (NQ and WebQA) used in the experiments.
\label{apendix:case_study}
\begin{table*}[p]
	\small
	\begin{tabular}{@{}p{\textwidth}@{}}
		\toprule
      \textbf{Question:} who got the first nobel prize in physics?  \\ 
      \textbf{Title:} Nobel Prize in Physics. \\ 
      \textbf{Passage:} receive a diploma, a medal and a document confirming the prize amount. Nobel Prize in Physics The Nobel Prize in Physics () is a yearly award given by the Royal Swedish Academy of Sciences for those who have made the most outstanding contributions for mankind in the field of physics. It is one of the five Nobel Prizes established by the will of Alfred Nobel in 1895 and awarded since 1901; the others being the Nobel Prize in Chemistry, Nobel Prize in Literature, Nobel Peace Prize, and Nobel Prize in Physiology or Medicine. The first Nobel Prize in Physics was. based on this passage. \\
      \textbf{UPR Question:} What is the Nobel Prize in Physics? \\              
	 \textbf{\model Question:} Who was the first to receive the Nobel Prize in Physics?
 \\
  \midrule   
	  \textbf{Question:} when is the next deadpool movie being released?  \\ 
       \textbf{Title:} Deadpool (film).\\ 
       \textbf{Passage:} Screen Rant called it possibly "the best film marketing campaign in the history of cinema". HostGator's Jeremy Jensen attributed the campaign's success to Reynolds, and to Fox for embracing the film's R rating. "Deadpool"s world premiere was held at the Grand Rex in Paris on February 8, 2016, before its initial theatrical release in Hong Kong the next day. This was followed by releases in 49 other markets over the next few days, including the United States on February 12. The movie was released in several formats, including IMAX, DLP, premium large formats, and D-Box. Kinberg explained that unlike the.\\ 
      \textbf{UPR Question:}  What was the first movie Deadpool was released in?\\              
	 \textbf{\model Question:} When was the world premiere of the film "Deadpool" held?\\
\midrule   
	  \textbf{Question:}  which mode is used for short wave broadcast service?  \\ 
    \textbf{Title:}  Sender Zehlendorf.\\ 
    \textbf{Passage:} Passage:  of Russia, partly in the Simulcast mode. The long wave transmitter changed over on 29 August 2005 as first German large transmitter to Digital Radio Mondiale. The long wave transmitter ceased operation on December 31, 2014 as part of the general shutdown in Germany of AM radio services to the public. The mast continues to support VHF radio antennas providing FM broadcast services. Sender Zehlendorf Sender Zehlendorf is a radio transmission facility which has been in service since 1936, when a short wave transmitter was built in Zehlendorf (a village near Oranienburg) as part of the establishment of permanent radio.\\
        \textbf{UPR Question:}  What is the name of the transmitter?\\   
        \textbf{\model Question:} What digital broadcasting technique did the long wave transmitter at Sender Zehlendorf switch to on August 29, 2005?
\\

\midrule   
	  \textbf{Question:}  the south west wind blows across nigeria between?  \\ 
    \textbf{Title:}   Oron people.\\ 
    \textbf{Passage:} Passage:  Civil War. Oron is found in the flood plain of South Eastern Nigeria, with the land mainly intersected by numerous streams and tributaries flowing into Cross River. The entire coastline stretches from Uya Oron to Udung Uko. Oron is in the tropical region and has a uniformly high temperature all the year round. The two main seasons are the dry which spans between October and April and wet season which starts around May and ends in September. There are also two prevailing winds – the South-West onshore winds which brings heavy rains and the North- East trade winds blowing across.\\
        \textbf{UPR Question:}  What is the name of the river that flows into Cross River?\\   
        \textbf{\model Question:} What are the main geographical and climatic features of the Oron region in South Eastern Nigeria?
\\
  \bottomrule             
	\end{tabular}
	\caption{ Examples of generated questions, Documents retrieved by BM25 for the Natural Questions (NQ) test dataset, comparing UPR and \model-based re-ranking. }
	\label{apendix:nq_case_study_retriever}
\end{table*}

\begin{table*}[p]
	\small
	\begin{tabular}{@{}p{\textwidth}@{}}
		\toprule
      \textbf{Question:} who plays ken barlow in coronation street? \\ 
      \textbf{Title:}   Ken Barlow.  \\ 
      \textbf{Passage:} Ken Barlow Kenneth Barlow is a fictional character from the British ITV soap opera \"Coronation Street\", played by William Roache. Ken was created by Tony Warren as one of \"Coronation Street's\" original characters. He debuted in the soap's first episode on 9 December 1960. Having appeared continuously since, December 2010 marked Ken's 50th anniversary on-screen, making Roache the longest-serving actor in a televised soap opera. Roache was honoured at the 2010 Guinness World Records ceremony for the achievement, when he surpassed actor Don Hastings from the American soap opera \"As the World Turns\", who previously held this title. Roache stated.\\
      \textbf{UPR Question:} Who is the longest-serving actor in a televised soap opera?\\              
	 \textbf{\model Question:} Who created the character Ken Barlow for "Coronation Street"?
 \\
  \midrule   
	      \textbf{Question:}   what does jamaican people speak?\\ 
      \textbf{Title:}   PhoneShop. \\ 
      \textbf{Passage:}  lives with his mother. He drives an Audi TT with numerous mechanical faults, and it is revealed that he paid "£3,000 for a £30,000 car." Jerwayne is Ashley's best friend as well as close co-worker, and the pair frequently egg each other on in foolish or unrealistic schemes and escapades. Jerwayne has an ongoing romantic interest in Janine, which is highlighted when Janine falls for a seemingly suave and successful Scotsman, provoking a jealous reaction from Jerwayne who vows to disrupt the relationship (which he eventually does with Ashley's help). Jerwayne speaks in a loud, Jamaican English dialect, is of.\\
      \textbf{UPR Question:} What is the name of Ashley's best friend? \\              
	 \textbf{\model Question:}  What kind of car does Jerwayne drive?
 \\
\midrule   
      \textbf{Question:}  what did james k polk do before he was president?\\ 
      \textbf{Title:}  Washington C. Whitthorne. \\ 
      \textbf{Passage:} Washington C. Whitthorne Washington Curran Whitthorne (April 19, 1825September 21, 1891) was a Tennessee attorney, Democratic politician, and an Adjutant General in the Confederate Army. Whitthorne was born near Petersburg, Tennessee in Marshall County. One day when Whitthorne was young James K. Polk stayed at his family's home. Polk saw how bright he was and asked, "What are you going to make of this boy?" His father replied "I am going to make him the President of the United States." Polk then told them to send the boy to Columbia and he would make him a lawyer. He attended Campbell. \\
      \textbf{UPR Question:} What was Washington Curran Whitthorne's profession?\\              
	 \textbf{\model Question:}  What job did Washington C. Whitthorne have before becoming the Adjutant General in the Confederate Army?

 \\
  \bottomrule             
	\end{tabular}
	\caption{ Examples of generated questions, Documents retrieved by BM25 for the WebQA test dataset, comparing UPR and \model-based re-ranking }
	\label{apendix:webq_case_study_retriever}
\end{table*}

\section{Major and Minor Categories}
\label{appendix:categories}

Table \ref{tab:major_minor_categories} presents the list of major and minor categories used for classification in our experiments. Each major category includes its associated minor types and their descriptions.

\begin{table*}[ht]
\centering
\small
\begin{tabular}{|l|l|l|}
\hline
\textbf{Major Category} & \textbf{Minor Category} & \textbf{Description} \\ \hline
\textbf{ABBREVIATION (ABBR)} & ABBR:abb (0) & Abbreviation \\ \cline{2-3}
                             & ABBR:exp (1) & Expression abbreviated \\ \hline
\textbf{ENTITY (ENTY)}       & ENTY:animal (2) & Animal \\ \cline{2-3}
                             & ENTY:body (3)   & Organ of body \\ \cline{2-3}
                             & ENTY:color (4)  & Color \\ \cline{2-3}
                             & ENTY:cremat (5) & Invention, book, and other creative piece \\ \cline{2-3}
                             & ENTY:currency (6) & Currency name \\ \cline{2-3}
                             & ENTY:dismed (7)  & Disease and medicine \\ \cline{2-3}
                             & ENTY:event (8)   & Event \\ \cline{2-3}
                             & ENTY:food (9)    & Food \\ \cline{2-3}
                             & ENTY:instru (10) & Musical instrument \\ \cline{2-3}
                             & ENTY:lang (11)   & Language \\ \cline{2-3}
                             & ENTY:letter (12) & Letter (a-z) \\ \cline{2-3}
                             & ENTY:other (13)  & Other entity \\ \cline{2-3}
                             & ENTY:plant (14)  & Plant \\ \cline{2-3}
                             & ENTY:product (15) & Product \\ \cline{2-3}
                             & ENTY:religion (16) & Religion \\ \cline{2-3}
                             & ENTY:sport (17)  & Sport \\ \cline{2-3}
                             & ENTY:substance (18) & Element and substance \\ \cline{2-3}
                             & ENTY:symbol (19) & Symbols and sign \\ \cline{2-3}
                             & ENTY:techmeth (20) & Techniques and method \\ \cline{2-3}
                             & ENTY:termeq (21) & Equivalent term \\ \cline{2-3}
                             & ENTY:veh (22)    & Vehicle \\ \cline{2-3}
                             & ENTY:word (23)   & Word with a special property \\ \hline
\textbf{HUMAN (HUM)}        & HUM:gr (28) & Group or organization of persons \\ \cline{2-3}
                             & HUM:ind (29) & Individual \\ \cline{2-3}
                             & HUM:title (30) & Title of a person \\ \cline{2-3}
                             & HUM:desc (31) & Description of a person \\ \hline
\textbf{LOCATION (LOC)}      & LOC:city (32)   & City \\ \cline{2-3}
                             & LOC:country (33) & Country \\ \cline{2-3}
                             & LOC:mount (34)   & Mountain \\ \cline{2-3}
                             & LOC:other (35)   & Other location \\ \cline{2-3}
                             & LOC:state (36)   & State \\ \hline
\textbf{NUMERIC (NUM)}       & NUM:code (37)  & Postcode or other code \\ \cline{2-3}
                             & NUM:count (38) & Number of something \\ \cline{2-3}
                             & NUM:date (39)  & Date \\ \cline{2-3}
                             & NUM:dist (40)  & Distance, linear measure \\ \cline{2-3}
                             & NUM:money (41) & Price \\ \cline{2-3}
                             & NUM:ord (42)   & Order, rank \\ \cline{2-3}
                             & NUM:other (43) & Other number \\ \cline{2-3}
                             & NUM:period (44) & Duration or time period \\ \cline{2-3}
                             & NUM:perc (45)  & Percentage, fraction \\ \cline{2-3}
                             & NUM:speed (46) & Speed \\ \cline{2-3}
                             & NUM:temp (47)  & Temperature \\ \cline{2-3}
                             & NUM:volsize (48) & Size, area, or volume \\ \cline{2-3}
                             & NUM:weight (49) & Weight \\ \hline
\end{tabular}
\caption{Major and Minor Categories used for classification in our experiments.}
\label{tab:major_minor_categories}
\end{table*}

\end{document}